\documentclass{sig-alternate}
\usepackage{etoolbox}
\makeatletter
\patchcmd{\maketitle}{\@copyrightspace}{}{}{}
\makeatother
\usepackage{amsmath}
\usepackage{multirow}
\usepackage{amssymb}
\DeclareMathOperator*{\argmax}{argmax}
\usepackage{graphicx}
\usepackage{subfigure}
\usepackage{epsfig}
\usepackage{balance}
\begin{document}


%

\title{Store Location Selection via Mining Search Query Logs of  Baidu Maps }

\newcommand{\map}{map query}
\newcommand{\wifi}{WiFi Connection}
\newcommand{\poi}{POI}
\newcommand{\para}[1]{{\vspace{3pt} \bf \noindent #1 \hspace{3pt}}}
\newcommand{\system}{\textsc{D3SP}}
\newcommand{\eat}[1]{}
\newcommand{\fixme}[1]{{\color{red} #1}}

\author{Mengwen Xu$^\S$$^\dagger$, Tianyi Wang$^\S$, Zhengwei Wu$^\S$, Jingbo Zhou$^\S$,Jian
  Li$^\dagger$, Haishan Wu$^\S$\titlenote{Haishan Wu is the corresponding author.}\\
\affaddr{$^\S$Baidu Research, Big Data Lab, $^\dagger$IIIS, Tsinghua University}\\
\{xumengwen,wangtianyi02,wuzhengwei,zhoujingbo,wuhaishan\}@baidu.com\\
lijian83@mail.tsinghua.edu.cn
}

\maketitle
\begin{abstract}
Choosing a good location when opening a new store is crucial for the
future success of a business. Traditional methods include offline
manual survey, which is very time consuming, and analytic models based
on census data, which are unable to adapt to the dynamic market. The
rapid increase of the availability of big data from various types of
mobile devices, such as online query data and offline positioning
data, provides us with the possibility to develop automatic and
accurate data-driven prediction models for business store
placement. In this paper, we  propose a Demand Distribution Driven
Store Placement (D3SP) framework for business store placement by
mining search query data from Baidu Maps. 
 \system\ first detects the spatial-temporal
distributions of customer demands on different business services via
query data from Baidu Maps, the largest online map search engine in
China, and detects the gaps between demand and supply. Then we
determine candidate locations via clustering such gaps. In the final
stage, we solve the location optimization problem by predicting
and ranking the number of customers. We not only deploy supervised
regression models to predict the number of customers, but also learn
to rank models to directly rank the locations.  
We
evaluate our framework on various types of businesses in real-world
cases, and the experiments results demonstrate the effectiveness of
our methods.  \system\  as the core function for store placement has
already been implemented as a core component of our business analytics platform and could be potentially used by chain store merchants on Baidu Nuomi.
\end{abstract}

%


\section{Introduction}
Selecting a location when opening a new store is crucial for future
business development: a good location may lead to very good business,
while an incorrect one may result in serious business risk and even
the failure of business.   

Many efforts have been made to address this task both qualitatively
and quantitatively. For example, one can conduct offline manual survey
on one potential location. And many business consulting companies
provide consulting services by collecting and mining data from
third parties such as census and demographic data from the
government. 
Some previous work~\cite{berman2002generalized, xiao2011optimal,
  chen2014efficient} study optimal location problem as a covering
problem which maximizes the number of people a location can attract.  

Nowadays, personal mobile devices are ubiquitous with large
scale mobile data, making it possible for us to offer big data
driven solutions for business solutions, especially for business location
selection in our topic here. In a recent work~\cite{karamshuk2013geo},
the authors have developed a data-driven approach to solve the problem
of retail store placement by predicting the check-in numbers at given
locations using linear supervised learning model. However, these
models do not capture the targeted user demands (i.e., demands for a
category of stores).  
\begin{figure}[t]
\centering
\subfigure[Detecting the spatial distribution of customer demand from query data from Baidu Maps.]{ 
\label{fig:intro:a} 
\includegraphics[width=0.22\textwidth]{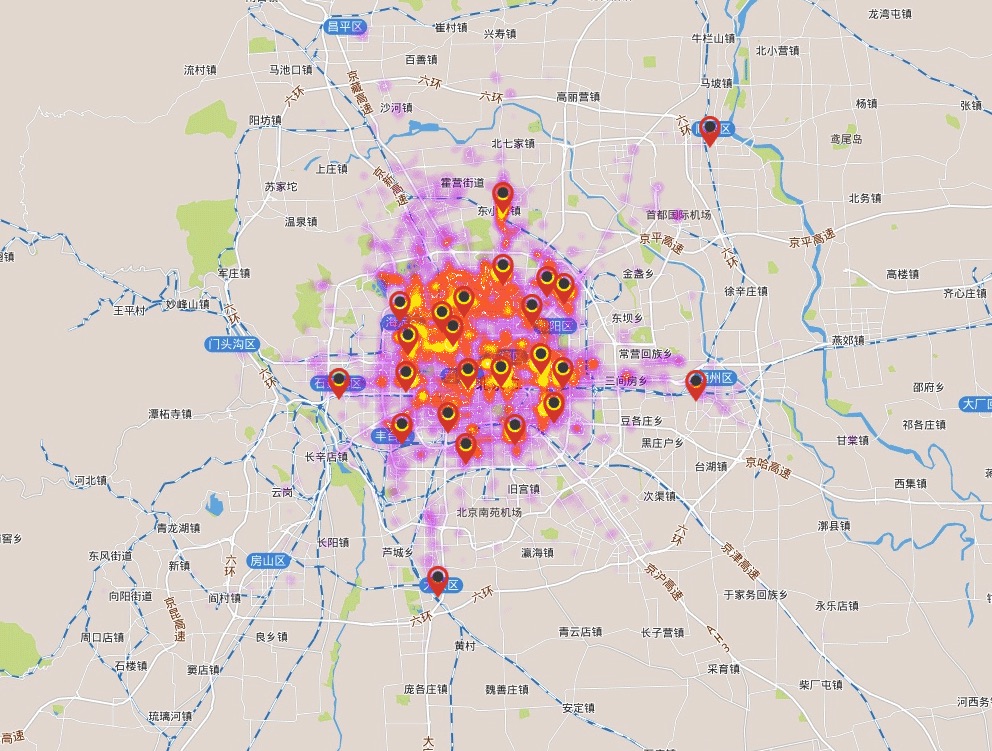}
} 
\subfigure[Estimating service distance of existing stores and detecting the demand-supply gap.]{ 
\label{fig:intro:b} 
\includegraphics[width=0.22\textwidth]{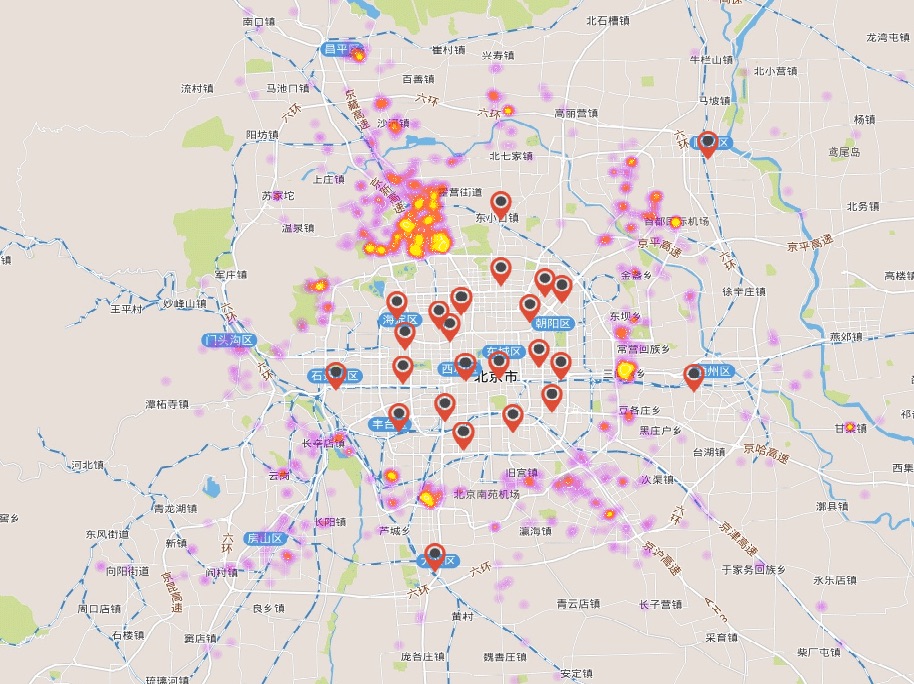}
} 
\subfigure[Determing potential placement locations by clustering the locations that have demand-supply gaps.]{ 
\label{fig:intro:c} 
\includegraphics[width=0.22\textwidth]{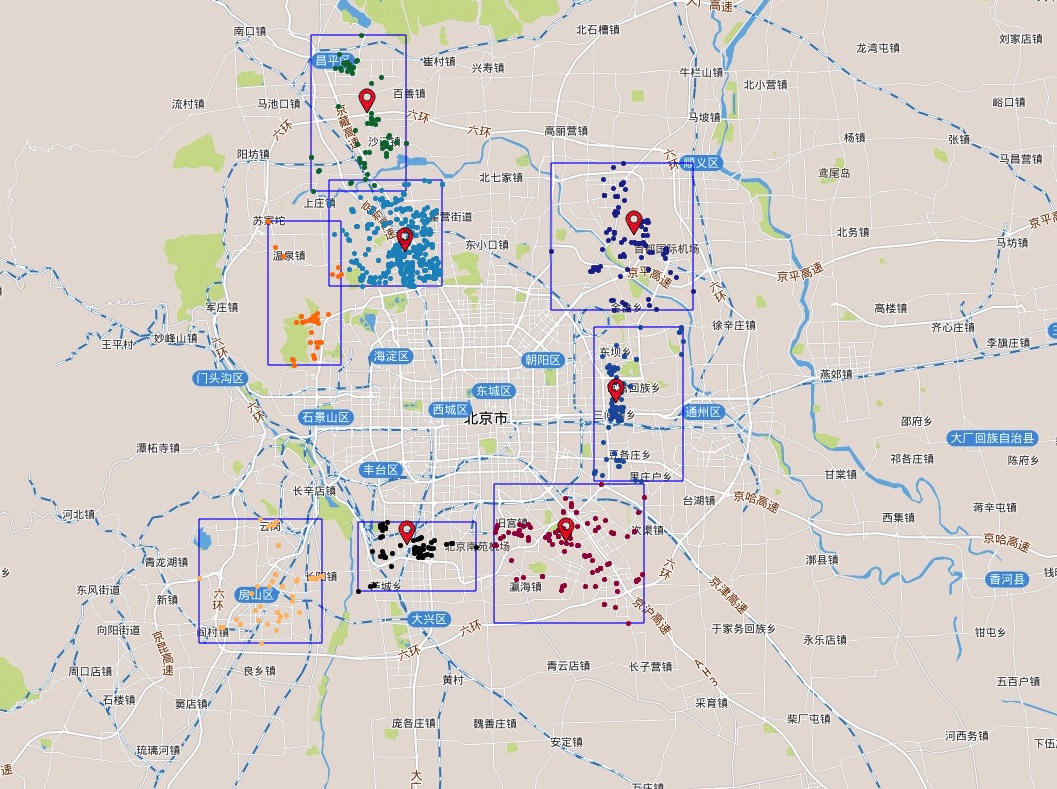}
} 
\subfigure[Optimizing the placement positions by maximizing the predicted visited customers.]{ 
\label{fig:intro:d} 
\includegraphics[width=0.22\textwidth]{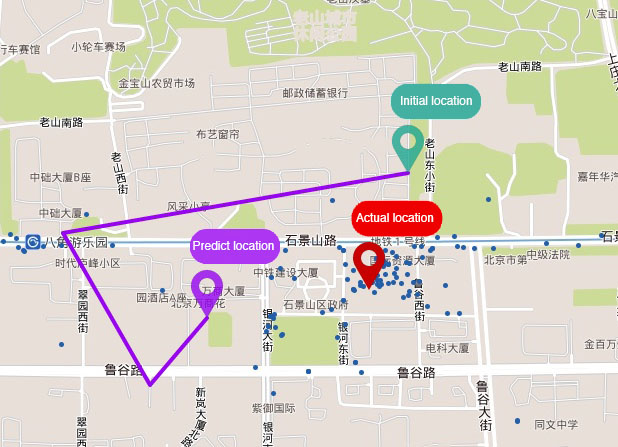}
} 
\caption{Example of finding optimal placement for opening a new ``Haidilao'' using our framework. Position marks are the existing stores, and the hot points are the locations where users query the store.}
\label{fig:intro}
\end{figure}

In this paper, we propose a Demand Distribution Driven Store
Placement(D3SP) framework, and utilize
 multiple spatial-temporal data
sources 
to accomplish this task. 
Given the category of a new store to be opened, the aim of our
approach is to identify several most promising locations that meet user
demands and attract a large number of potential customers. In this
paper, we take advantage of search query data from Baidu Maps and integrate
other types of spatial-temporal data sources such as POI data etc 
to solve
the problem.  

Location based applications like map search and navigation systems
have been developed in recent years. The will of users going to some
places can be inferred from this kind of data (i.e., map query data), 
 demonstrating the potential consumptive demands of users~\cite{wupredict}. 
Generally, the strategy of finding optimal placement
is to find the most suitable locations from a list of candidate
locations~\cite{wood2012leveraging}. Here, we define those candidate
locations as the places where demands exceed supplies. An example of
finding optimal placement for opening a new ``Haidilao'' using
\system\ is shown in Figure~\ref{fig:intro}. For example, when users
query ``Haidilao'' (a popular hot pot restaurant in China), their
locations are recorded. In Figure~\ref{fig:intro:b}, after excluding
the locations within 5km around the existing stores, there are still
many places that do not meet user demands.  It is possible to cluster
the gaps between demand and supply (Figure~\ref{fig:intro:c}), and
then optimize the placement to obtain most customers (Figure~\ref{fig:intro:d}). 


Location placement is naturally a ranking problem. There are at least
two methods to achieve. We can first
predict the customer numbers and then rank manually, or utilize learn
to rank methods directly. In this paper, we will try both and evaluate
the results with real-world cases.  A lot of factors can impact the
number of customers a location can attract.
We consider 
economic development, popularity and transportation convenience
~\cite{rogers2007retail} in our model.  The models are trained and
validated with existing stores' locations and customers' behavior.

Our work makes the following contributions: 
\begin{itemize}
 \item  First, we identify the spatial distribution of user demands from map
   query data, and obtain the places that demands exceed supplies. 
 Here, we extracted two types of demands of real users: specific
 demands (for specific brand of chain store) and general demands (for
 a general category) from the data.  
\item Second, we propose practical features and models to rank the candidate
  locations. We utilize both supervised regression models to predict
  the number of customers, and learn-to-rank models to rank the
  candidate locations directly. Novel features such as
    distance to the center of the city, popularity of specific area
    category and real estate price nearby are also proposed. 
\item Finally, we evaluate our methods with real-world cases, and developed a completed system to find optimal placement
  for opening a new store. 
\system\ is a crucial part of the
  location-based business support system developed in Baidu. We
  evaluate \system\ on both experiments and the real cases. The
  experiments show that our framework predicts accurately for various
  categories, and the real cases show that our framework works well in
  the real world.  
\end{itemize}
\section{Data and Framework}

\subsection{Data and Preprocessing}
\label{sec:datapreprocess}
In this section, we provide details of the heterogeneous data sets we
use in this paper, as well as the preprocessing of these data
sets. The data sets include \map\ data, \poi\ data and \wifi\ data.

\para{Location query data from Baidu Maps.}
Map query data
contains over one billion queries from 500 million users using Baidu Maps on mobile phones.  
Each \map\ record has the following information: an anonymized user
ID, timestamp, latitude and longitude of the location where the query
was issued, query keywords with the associated POI attributes.

When a user searches for a place or navigates a route to a place using
Baidu Maps,  the query action reflects a demand of a specific
place or a category. 
The actual location where the user issues a query also implies a spatial demand.   
For example, a query [$u_i$, 2015-08-08 08:42:28, (116.34, 40.02),
``Starbucks''] 
means that the user $u_i$ wants to have coffee at location (116.34,
40.02). Further analysis can be found in Section \ref{sec:lwd}.  

\para{\poi\ data.}
Point-of-Interest(POI) in a city is tightly related to people's
consumptive behavior. We can learn a lot from existing POIs that are
similar to a new store to be opened.
The \poi\ data set is a
collection of POIs. For each POI $p$, its attributes include POI ID
$p.id$, location $p.l$, or $p.x$ and $p.y$, category $p.c$ and other detailed
information. Attribute ``category'' has two hierarchical levels. For
example, category ``hotel'' is the upper level, and ``express inn'' is
the subcategory of ``hotel''. 

\para{\wifi\ data.}
We obtained \wifi\ data from a
  third-party data source,
which contains records of 3-month historical WiFi
connections
from June 1, 2015 to August 30,
2015. Each record contains an anonymized user id, timestamp, 
and the corresponding POI. 

\wifi\ data can be a good indicator of customer numbers, and serves as
part of ground truth in prediction of customer numbers.
It has
several advantages over the check-in data used in 
previous work~\cite{karamshuk2013geo}. First, only the customers can
connect the WiFi in the POI, which eliminates the situation of cheated
check-ins. Meanwhile, \wifi\ data has usually many more records than the check-in
data for a single user. A user often connects to wifi in restaurants
or hotels, but does not necessarily check in her locations.


   



\para{Integrating the data sets.}
By integrating  \wifi\ and  \poi, we obtain the total number of customers
for each POI. Note that each user can generate several records in one day at a
specific POI, and some POI may have several accessing points. However,
for a POI, the number of customers in one day should be the appearance
of all the users in that day. Therefore, we clean the data by
maintaining one record for a user in one day at a specific POI. 
Each record contains
a POI id, POI category, POI name, POI location and POI
customer number. In further analysis, we use the POI customer
number to measure how good a store location is. 

\para{Data privacy.} In this paper we were very careful to protect the
user privacy. We have set several key steps and protocols to protect the
privacy:
\begin{itemize}
\item First, all ids in our data are hashed and anonymous. None can be linked to
  any real-world identifier of a person. So we do not know the offline behavior of
  any real person.
\item Second, we have a very strict agreement with the third party
  that the \wifi\ data is kept secure, safe and anonymous.  
Each id is encrypted, and is not even linkable to the
  ids in map query data.
\item Finally, we are focused on analysis in aggregation level. For
  example, we aim to predict the number of customers of a location as
  a whole, instead of predicting whether a specific user will
  visit. None individual level analysis is deployed in our paper.
\end{itemize}
\subsection{Framework and methodology}
\label{sec:framework}
\begin{figure}[t]
\centering
\includegraphics[width=0.47\textwidth]{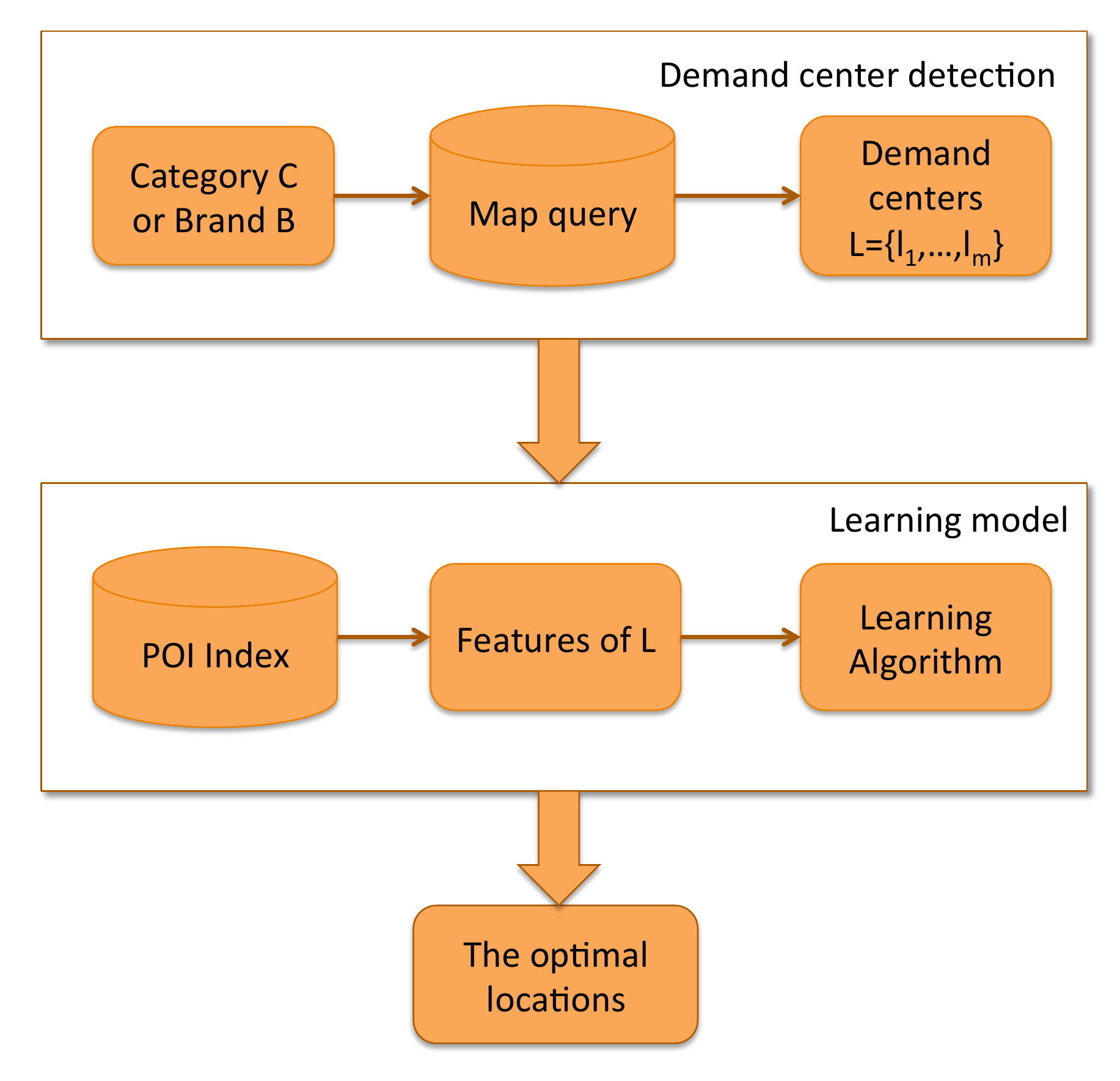}
\caption{Overview of demand distribution driven store placement
  framework. We first obtain several candidate locations by analyzing
  user demands. Then a supervised machine learning model is trained to
rank these candidate location.}
\label{fig:framework}
\end{figure}

A promising location for a new store has two requirements. First, it
should meet users' demands while there are not too many competitions. Second, it
needs to attract a significant number of customers. 
In this paper, we aim to first figure out a few
candidate locations by detecting demand centers, and then rank them
by predicting the custom number that each location may
have. Figure~\ref{fig:framework} shows the overall framework of our
paper, and more details are as follows.

In a city, there can be many locations to choose for opening a
store. Our first step is to
find several demand centers $L_d=\{l_1, l_2, ..., l_m\}$ as candidate locations for a store
category $C$ or a specific brand of chain stores $B$. Map query
data is used to find the demand points where the users query a POI or
a POI belonging to the same category. Thus, we obtain the demand
centers $L_d={l_1, \cdots, l_m}$ from those locations.

Then, we
transform the problem into a global ranking problem. Our 
goal is to rank the locations in $L_d$ for
opening a new store belonging to category. we achieve this by
predicting the customer number for each location in $L_d$. Here we
train a supervised machine learning model with the customer
numbers of existing stores. The features are mined from an area
$A_l$.  Here $A_l$ is a disc centered at $l$ with radius $r$. Considering
users' walking distance and GPS position shifting, we set $r$ to be 1
km. With this model, we are able to predict
the custom numbers of the candidate demand centers $L_d$ will have.
 Locations with higher values are more desirable to open new stores. 

\section{Obtaining demand centers}
\label{sec:lwd}
Our work is inspired by the fact that map queries reflect users' targeted
demands. Each query contains a source place and a
destination. The source place has demands while the destinations can
supply the demands. 
An ideal place for a new store to be opened is where there are many demands but few
supplies. In this section, we aim to identify such places with the
knowledge of map query data. 
We start from the analysis of user demands, and then
propose methods to identify the demand centers.
\begin{figure}[t]
\centering
\subfigure[Dynamic patterns of map queries and visitations.]{ 
\label{fig:map_consu:a} 
\includegraphics[width=0.4\textwidth]{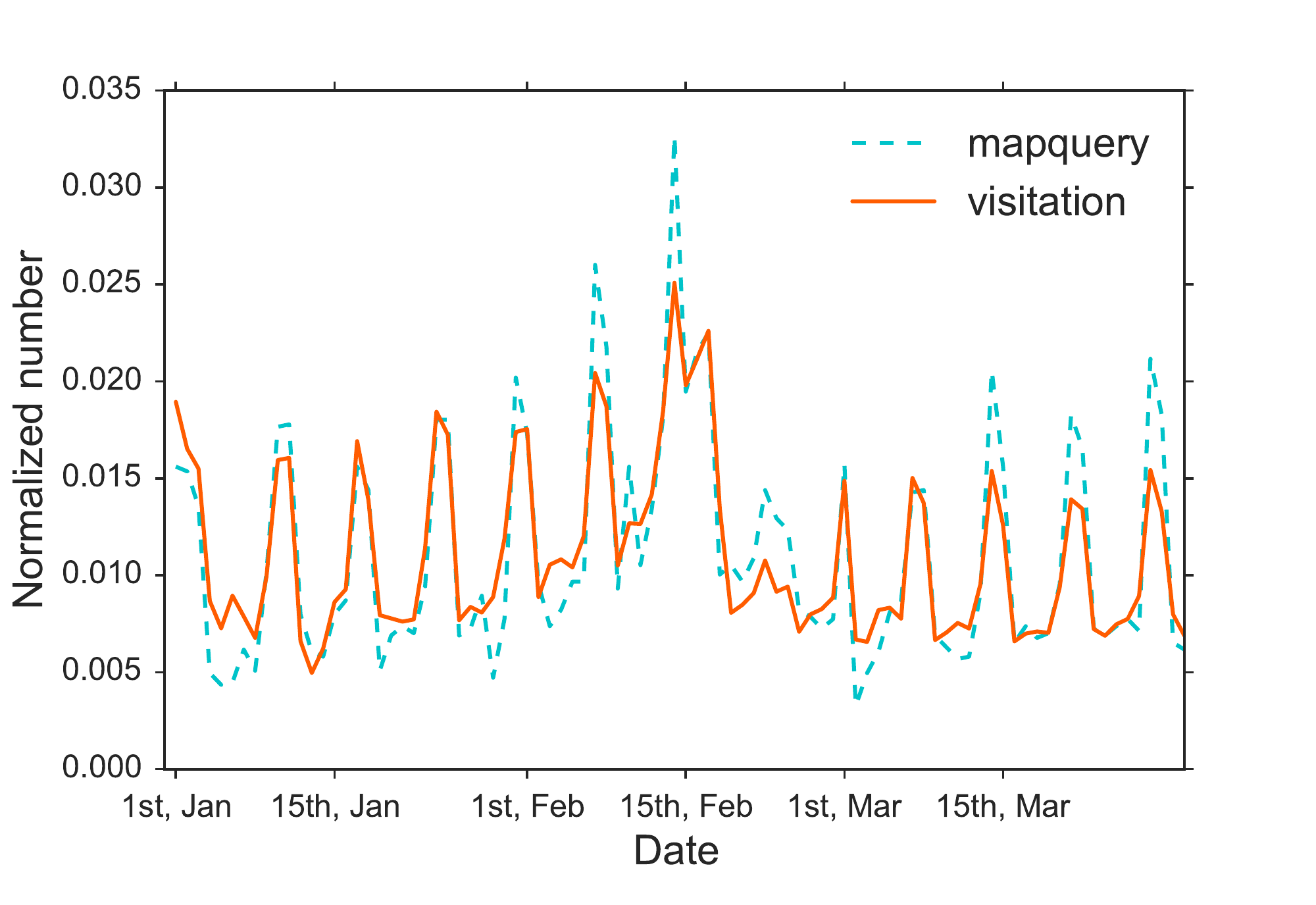}
} 
\centering
\subfigure[Coefficient of determination.]{ 
\label{fig:map_consu:b} 
\includegraphics[width=0.38\textwidth]{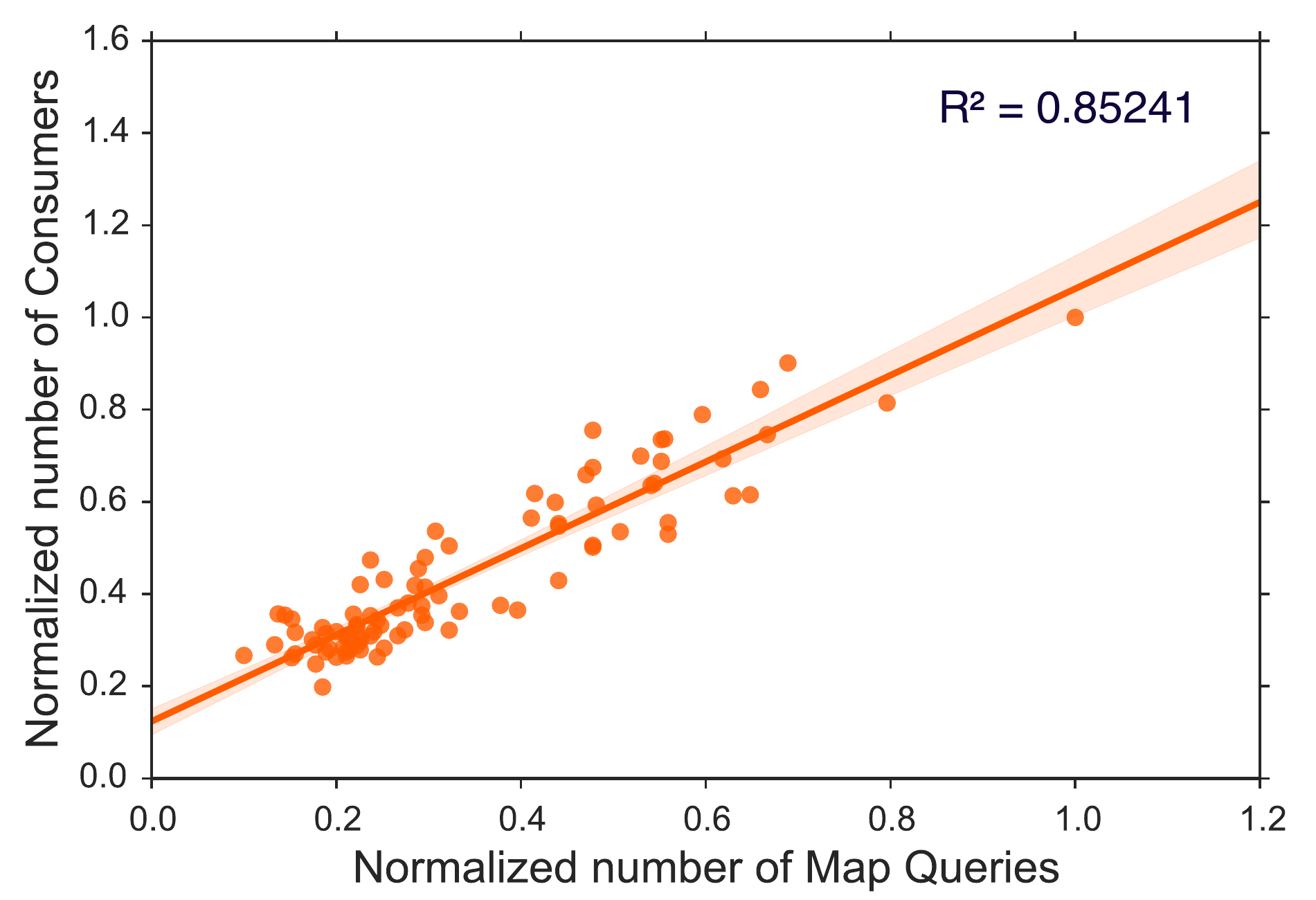}
} 
\caption{The comparison of map queries and visitations of a famous supermarket
  ``Wal-Mart''. (a) shows the normalized daily number of map queries and visitations during three months. (b) shows the high correlation of the number of map queries and visitations which means the number of map queries can be viewed as
  indicators of user demands.}  
  \label{fig:map_consu}
\end{figure}
\subsection{User Demand Analysis}

\begin{figure*}[t]
\begin{minipage}[t]{0.32\textwidth}
  \centering
  \includegraphics[width=1\textwidth]{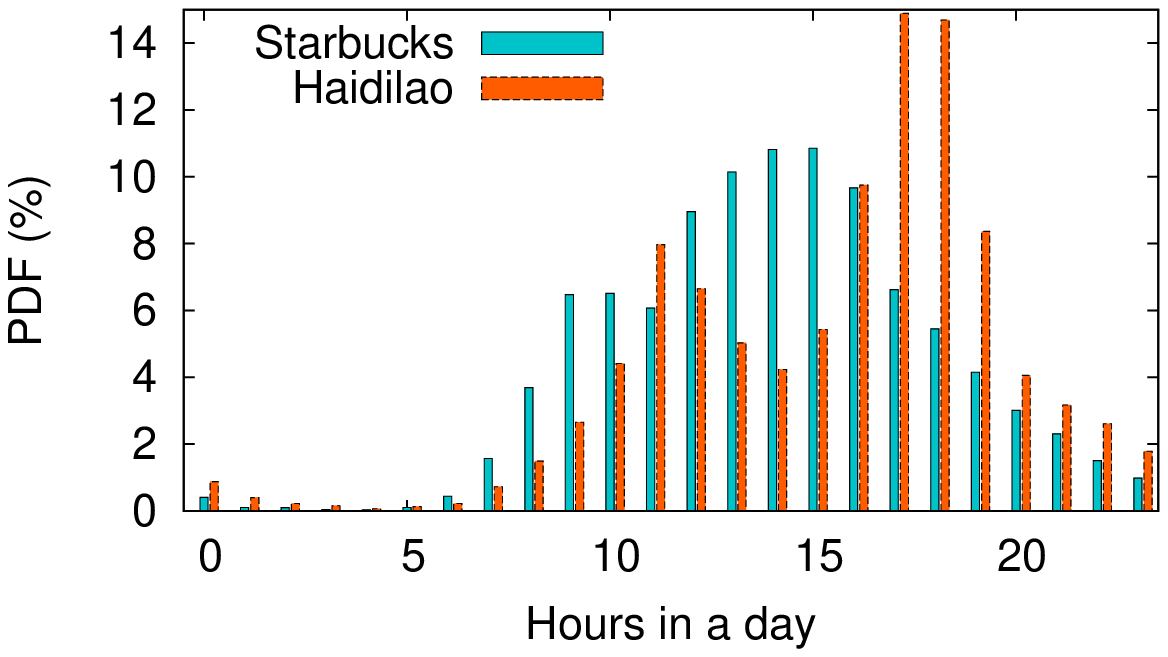}
  \caption{Distribution of demands through a day. Haidilao is biased
    to lunch and dinner time, while Starbucks attract visitation
    in the afternoon.}  \label{fig:hist_hour}
  \end{minipage}
\hfill
  \begin{minipage}[t]{0.32\textwidth}
  \centering
  \includegraphics[width=1\textwidth]{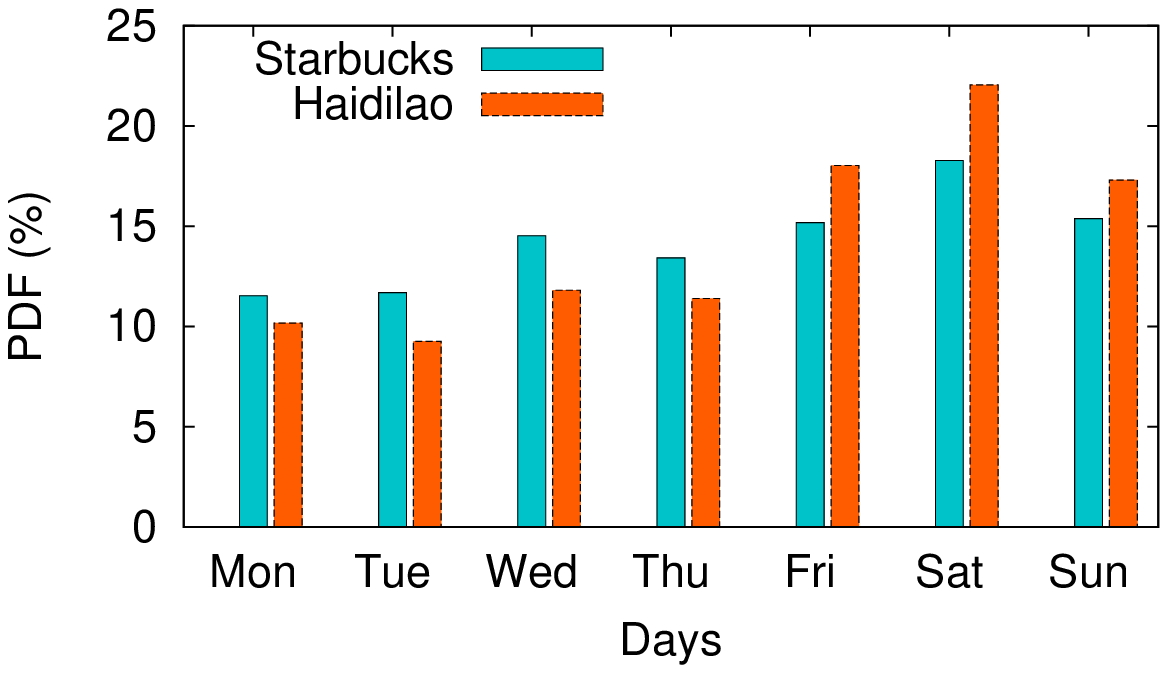}
  \caption{Distribution of demands through a week. Haidilao will
    attract many more customers during weekends.}  \label{fig:hist_day}
  \end{minipage}
\hfill
 \begin{minipage}[t]{0.32\textwidth}
  \centering
  \includegraphics[width=1\textwidth]{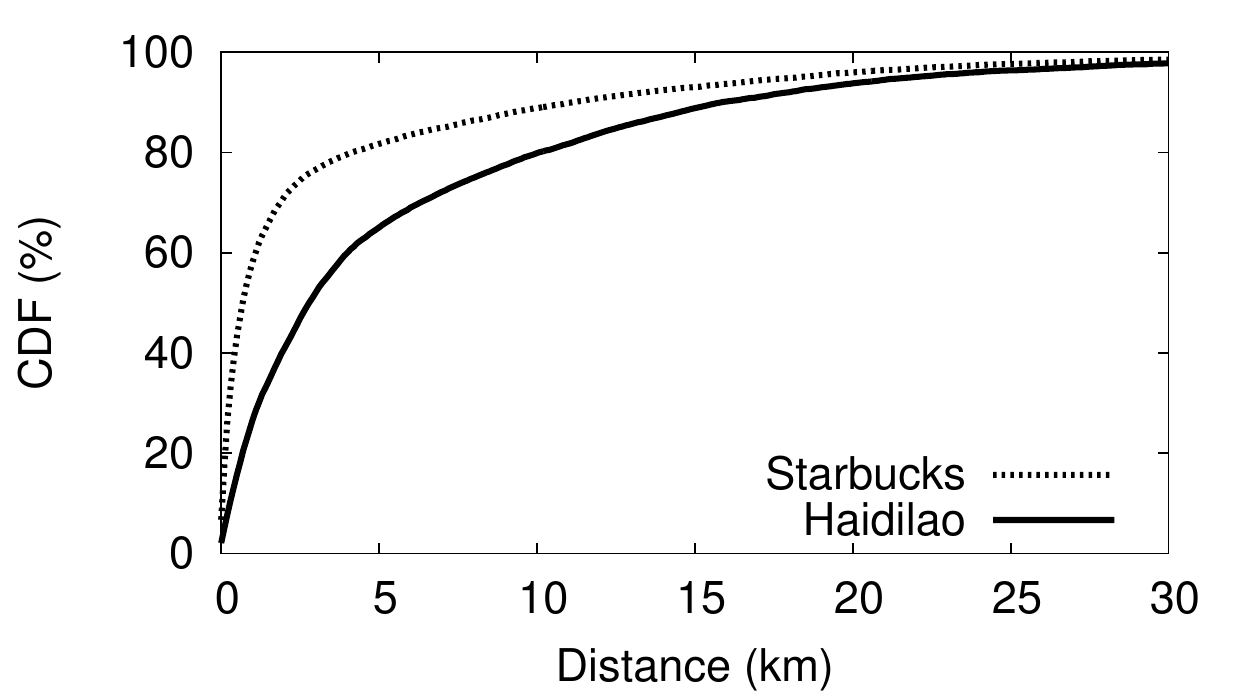}
  \caption{Cumulative distribution of source-destination distance for
    map queries. Users searching for Haidilao usually have larger
    travel distances. }  \label{fig:cdf_dist}
  \end{minipage}
\end{figure*}
An important part of our method is the modeling of user
demands. We will first demonstrate that map query reflects user
demands, and it is highly correlated with users' real visitation. 
We analyze the percentage of the
query sessions which users actually go to that place over all the
query sessions. 
We utilize the method in~\cite{wupredict} which assumes that a queried location is actually visited by the user if
there is at least one location record in user's next 1.5-day mobility
trace, and the distance of the location record and the queried location is less than 1 km. 
As
shown in Figure~\ref{fig:map_consu}, the normalized number of map
query and real visitations have similar dynamic patterns in a three
months period. In Figure~\ref{fig:map_consu:b} we also plot the coefficient
of determination $R^2$. The numbers of map queries and real visitations
have a correlation up to 0.85. These results confirm that map queries
are highly correlated with real visitations, and the number of map
queries can be indicators of customer numbers and real user demands.
More specifically, the visitation ratio can reach the ratio nearly 50\%, which is also a
strong evidence that there is a relationship between map queries and user demands.

Next, we will further analyze user demands via map query data. Remind that a map query record has the following attributes: timestamp
$T$, latitude $lat$ and longitude $lng$ of the location where the
query was issued, query keyword, attributes of POI. 
Formally, we denote a
demand point as $D=(lat, lng, t)$. That is, a demand point contains the locations
and a timestamp where and when the demand happens.  
We extracted two types of demands of real users. The first type is
{\em specific} demands. This type of demands can be extracted from
the route data in Baidu Maps. For example, a user has a specific need that he
wants to get to Starbucks. He or she searches for the
routes from his or her current location $(lat,lng)$ to a store of Starbucks
at time $t$. This implies that the user has a {\em specific} demand
$(lat,lng,t)$ on Starbucks. 
Another type of demands are {\em general} demands. An example is that
a user has a {\em general} demand that he wants to drink some
coffee, but he does not have preference on brands. This type of
demands can be extracted from the ``nearby'' queries which return POIs nearby according to the keyword.

After we define both specific and general demands, we conduct analysis
on the demand points we extract from queries. Our analysis basically
identifies the temporal and spatial differences among different
queries. Here we take Starbucks and Haidilao (a popular hotpot restaurant in China) as examples. 

\para{Active time.} In Figure~\ref{fig:hist_hour}
and~\ref{fig:hist_day} we plot the active time of the demand points. We find
different demands are active at different time. For example, people
usually get to Starbucks during the day, while they go to Haidilao at
noon and evening for meals. In a week, people prefer to go
to Haidilao on Friday, Saturday and Sunday, while people do not show
obvious preference on Starbucks. 

\para{Effective distance.} We also realize that people may tolerate
different distances for different stores. In
Figure~\ref{fig:cdf_dist}, we plot the distribution of distance between
source and destination. While 80\% queries of Starbucks are within
2 km, up to 40\% queries of Haidilao are more than 5 km. This is
understandable, since few people will travel 5 km just for a
coffee. Therefore, the nearby area of the location where the user
makes a query can be seen as a place where the user demand for the
queried place. 

The above analysis reveals that user demands may have various features
and patterns. When we further detect the demand centers, we need to
take the type of demands into consideration.

\subsection{Finding Demand centers}
To find demand centers, we have three
steps: identifying demand points, excluding supplies, and clustering demands.

\para{Identify demand points.} In the first step, we will identify and characterize the
demands, {\em e.g.}, Starbucks. As we explain before, we extract timestamp and location from
each query, including specific demands with targeted keywords Starbucks or
general demands with categories or tags of coffee from ``nearby'' queries.  We then characterize the
temporal and spatial features of demands.

\para{Exclude supplies.} After we identify the geographic distribution of
demand points, we need to identify the gap between demands and
supplies. 
For specific demands, we directly exclude areas where there are already
supplies. Here the exclusion is based on distance features. From the
distribution of source and destination distance, we identify a
threshold of 80\%, {\em i.e.} 80\% of users will not get to a store
farther from corresponding distance. Then for each existing store, we consider demand points
within the threshold are supplied. Thus these demand points are excluded. The remaining
demand points are the demand-supply gap points where demands exceed
supplies. For example in Figure~\ref{fig:cdf_dist}, 80\% of the users
search ``Starbucks'' are within distance 2km. The demand points within
2km from the existing stores are excluded. 

\begin{figure}[t]
\centering
\subfigure[Demand points and existing stores]{ 
\label{fig:demandcenter:a} 
\includegraphics[width=0.22\textwidth]{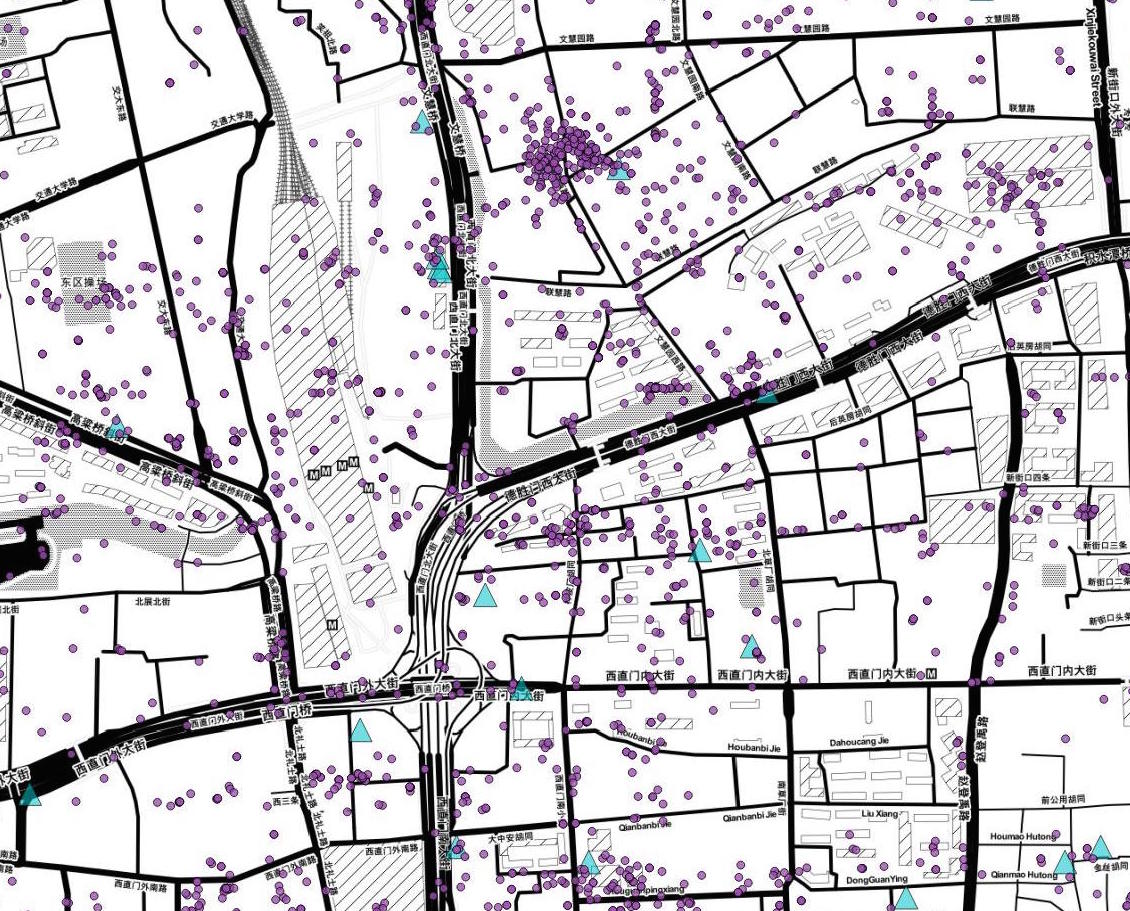}
} 
\subfigure[Remaining demand points and demand center]{ 
\label{fig:demandcenter:b} 
\includegraphics[width=0.2\textwidth]{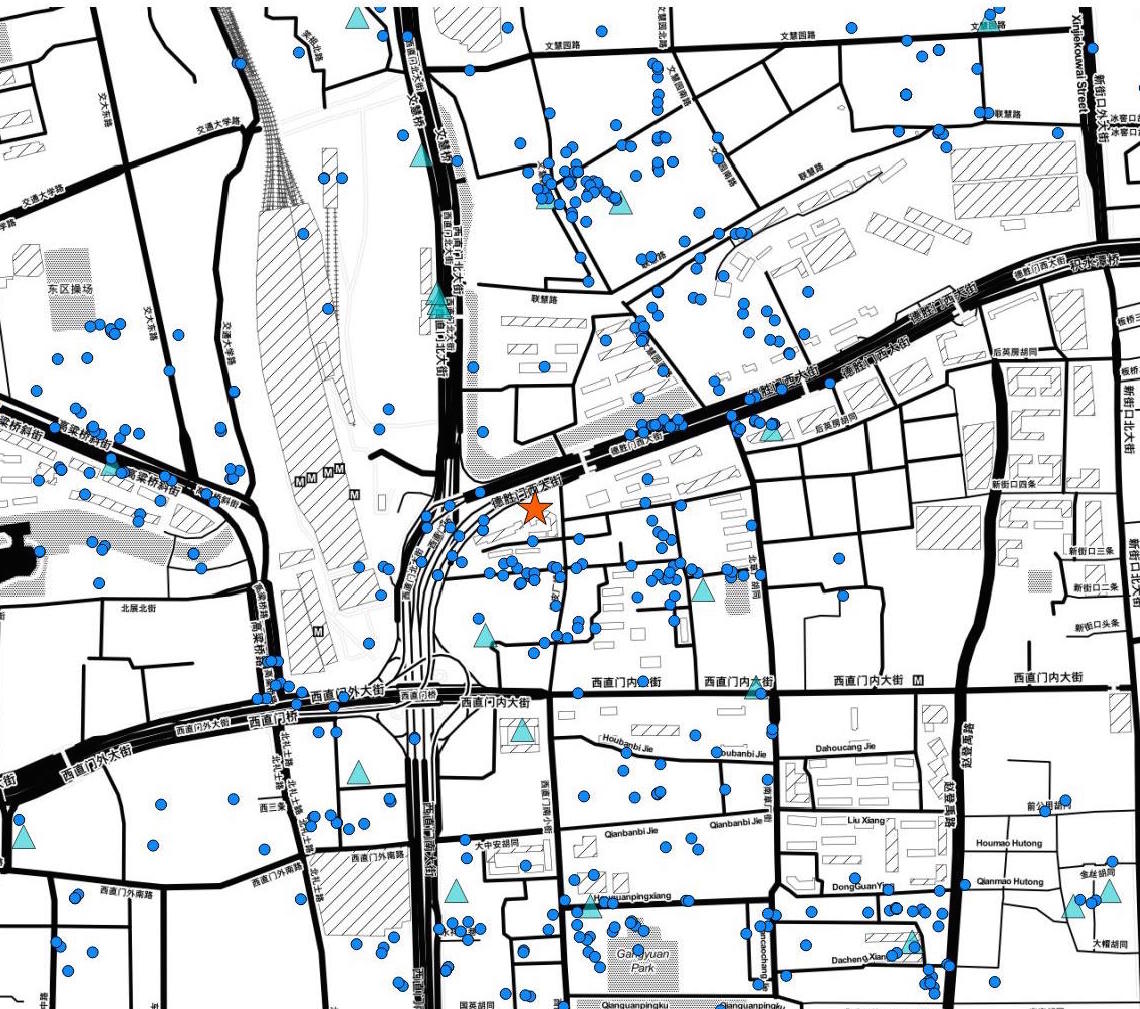}
} 
\caption{An example of exclude supplies with probability in the general demands case.}
\label{fig:demandcenter}
\end{figure}

For general demands, we exclude supplies with some probabilities,
because the user has the potential to a new store. 
We calculate the distance $d(l_u,l_n)$ between the nearest existing
store and the demand location $l_u=(lat,lng)$ of user $u$. 
Also the number of existing stores nearby is defined as $N$.
For a user $u$, if $d(l_u, l_n)$ is large, the probability of the
demand to be satisfied is low. On the other hand, if $N$ is large,  
the probability is high.
Therefore, we define two scores according to the two intuitions. One
is the distance score $S_d=1-e^{-d(l_u, l_n)^2/\sigma ^2}$, another is
the supply score $S_s=e^{-\epsilon N}$.  
The score of a demand to be remained $S_r=\alpha S_d + (1-\alpha)
S_s$. Here we set $\sigma =300$, $\epsilon =0.5$ and $\alpha=0.7$. 
Figure \ref{fig:demandcenter:a} shows the demand points of coffee
shops and existing stores in an area. Figure \ref{fig:demandcenter:b}
shows the remaining demand points after we remove the possible
satisfied ones. We can see that the demand points near the existng
stores are more likely to be excluded.

\para{Clustering demands.} After we identify the demand-supply gap points,
obtaining a place for a new store is still not trivial. This is because
the demand points are placed in space without any regularized
shapes. Here, we leverage machine learning clustering algorithms, and
select the cluster centers. Since we do not
know the exact number of clusters we will have, and would require a
center point of each cluster, we use
MeanShift~\cite{comaniciu2002mean} as our algorithm. We set the bandwidth parameter
as the distance threshold we identify in the previous step. This algorithm automatically
clusters the demand-supply gap points and returns a center for each cluster. we
denote the center points as demand centers which are candidate locations for store
placement. The star in Figure \ref{fig:demandcenter:b} is the demand center we find.

\section{Ranking promising locations}
With the demand centers, we next rank these promising locations. In
this paper we utilize supervised machine learning models to rank the
locations. There are two directions. First, we can directly apply
learn-to-rank methods. Second, we can predict the number of customers
with regression models, and then rank based on the predicted customer
numbers. We consider both in our paper, as illustrated in
Sec~\ref{sec:rank_alg}. Then we will introduce the features we use in
our models.

\subsection{Ranking Algorithms}
\label{sec:rank_alg}
Given a list of demand centers $L_m$, we utilize the supervised
learning methods to rank or predict the number of consumers of a new store
located at $l \in L_m$.  The locations with higher scores are the
optimal placements.  

\para{Baseline.}
The baseline method is to rank $L_m$ randomly, and the locations rank
at top are the results. 

\para{Learn to rank.}We first directly apply learn to rank
  methods. The method we use in this paper is LambdaMART~\cite{burges2010ranknet},
  which is a widely used learn to rank method in many works
  and the winner of 2010 Yahoo! Learning To Rank Challenge. It is a
  combination of 
  boosted tree MART and a learn to rank method LambdaRank. 

\para{Regression.}
For regression method, we consider the linear regression with regularization
(Lasso) and the kernel methods including Support Vector Regression
(SVR)~\cite{hearst1998support} and Kernel ridge regression
(KRR)~\cite{murphy2012machine}. We also consider two ensemble methods
Random Forests (RF) and Gradient Boosting Decision Tree (GBDT).  
 
Mathematically, Lasso consists of a linear model trained with L1 prior
as regularizer. The objective function to minimize is: 
\begin{equation}
\min_{w} { \frac{1}{2n_{samples}} ||X w - y||_2 ^ 2 + \alpha ||w||_1}
\end{equation}
where $X$ is feature vectors and $y$ is the target score. $\alpha$ is
a constant and $||w||_1$ is the $\ell_1$-norm of the parameter
vector. 

In SVR, training vectors are implicitly mapped into a higher (maybe
infinite) dimensional space by the kernel function. KRR combines Ridge
Regression (linear least squares with l2-norm regularization) with the
kernel trick. It thus learns a linear function in the space induced by
the respective kernel and the data. For non-linear kernels, this
corresponds to a non-linear function in the original space. 

Random Forests\cite{breiman2001random} and Gradient Boosting Decision
Trees\cite{friedman2001greedy} are two popular algorithms of the
ensemble method. 
Random Forests is a notion of the general technique of random decision
forests. It constructs a multitude of decision trees at training time,
and outputs the class that is the mean prediction of the individual
tree.  
Each tree in the ensemble is built from a sample drawn with
replacement from the training set.  

Gradient Boosting is typically used with decision trees of a fixed
size as base learners. For this special case Friedman proposes a
modification to Gradient Boosting method which improves the quality of
fit of each base learner.

\begin{figure}[t]
\centering
\subfigure[Heatmap of number of customers for coffee shops in Beijing]{ 
\label{fig:heatmap:a} 
\includegraphics[width=0.22\textwidth]{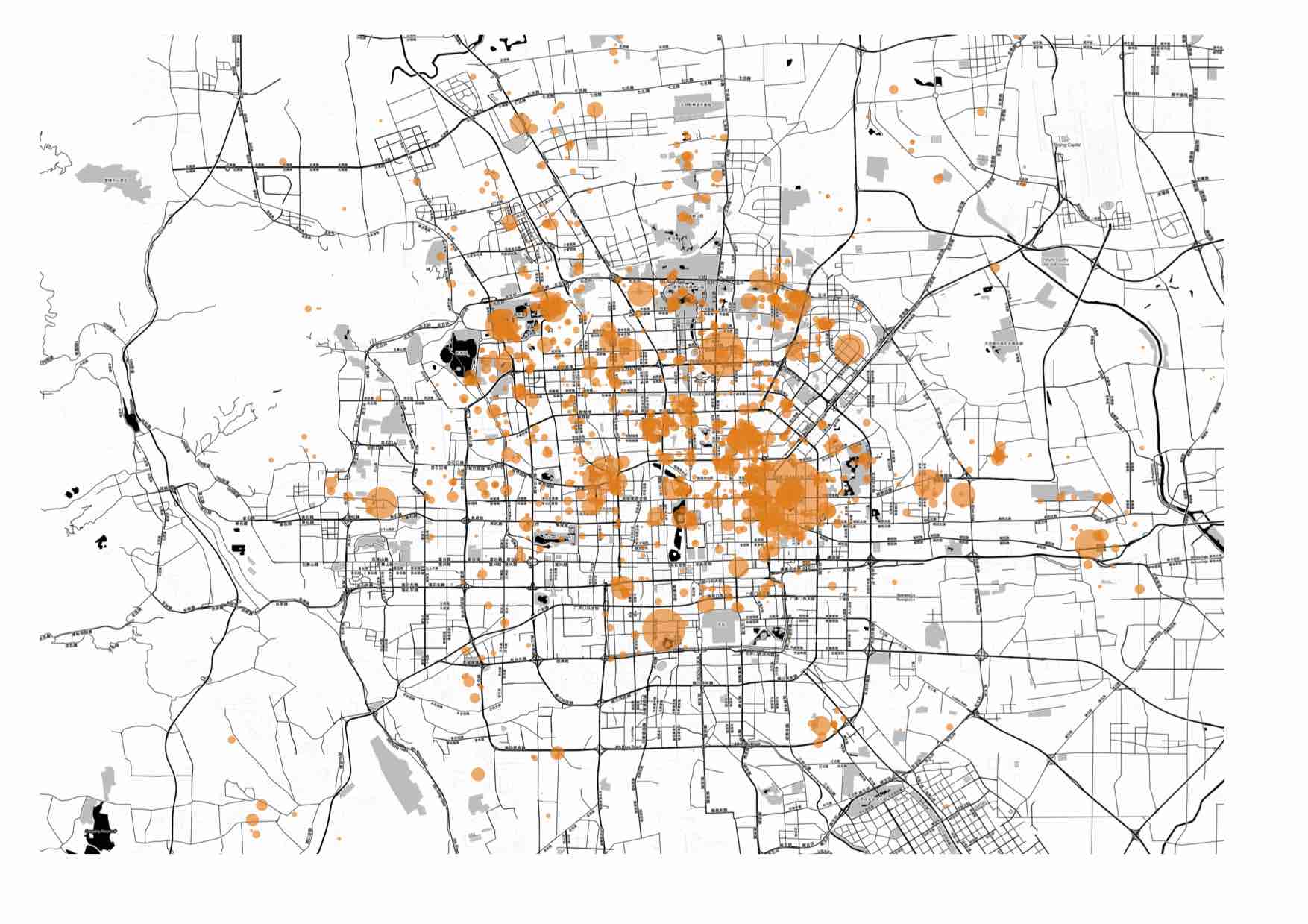}
} 
\subfigure[Heatmap of number of customers for Express inns in Beijing]{ 
\label{fig:heatmap:a} 
\includegraphics[width=0.22\textwidth]{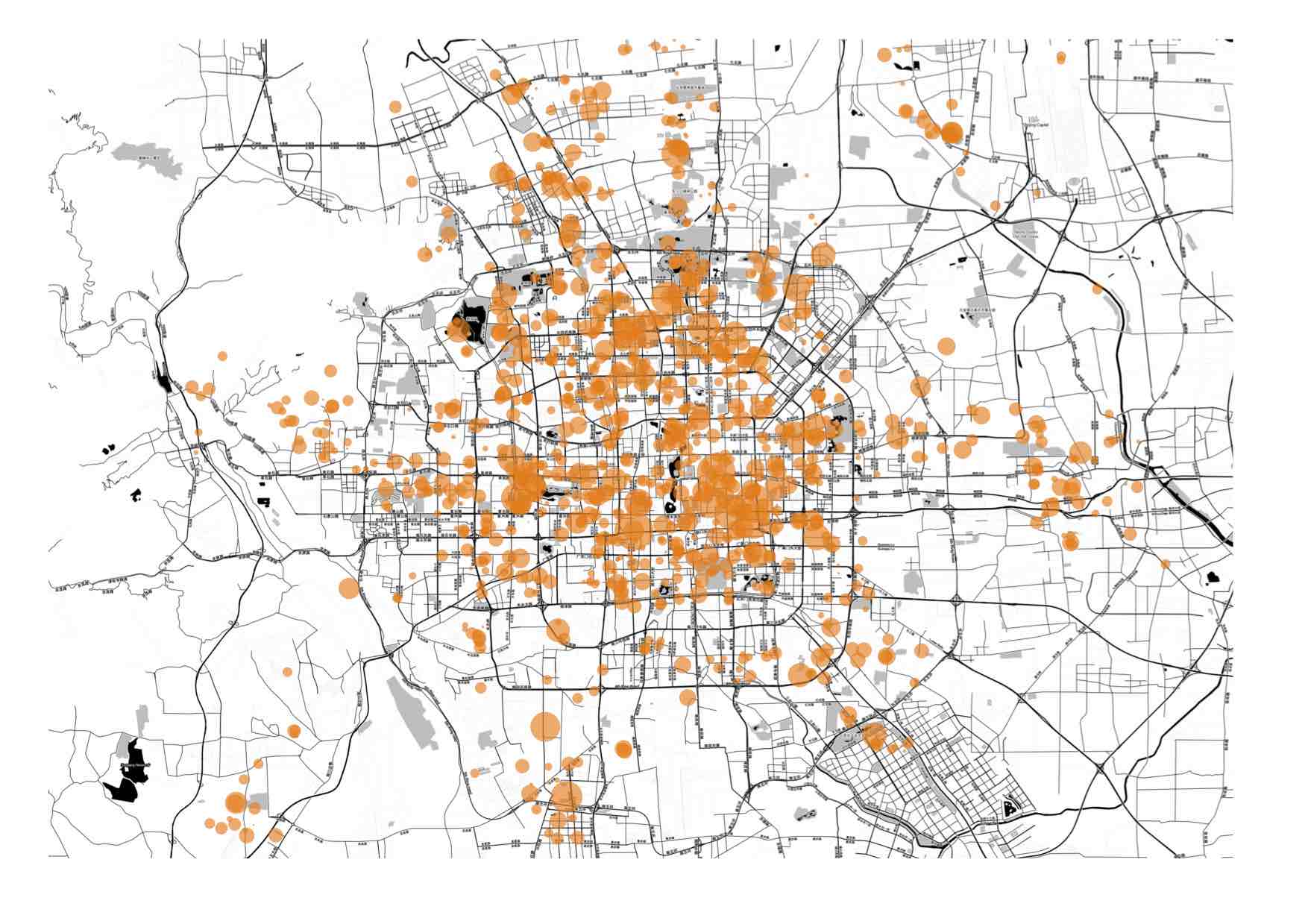}
} 
\caption{Geographic distribution of customers for coffee shops and express inns. The reder color means there are more customers. }
\label{fig:heatmap}
\end{figure}

\subsection{Features}
In this part, we will introduce the features mined from the multi-source data. 
Those features can reflect the quality of the area for the new
store including functionality, transportation convenience,
competition, economic development and so on. When given a location $l$
and the category $C$ of the new store to be opened, the features $F_l$
can be mined from the integrative dataset which is indexed in the
geographic POI index. Remind that the Area around $l$ is $A_l$ which
is a disc centered at $l$ with radius $r$ (1 km).  

\noindent{\textbf{Distance to the center of the city.}}
The popularity of a store is not only related with the local
information about the location, but also related with its geographic
position in a city.  To see how the customers distribute for a kind of
stores in a city, we draw the heat maps of the consumption behaviors
for express inns and coffee shops in Beijing, respectively in Figure
\ref{fig:heatmap}. The hotter(red color) the color of a point is, the
more customers there are. We find locations near the center of a city
usually attract more customers.

From the geographic aspect, we consider the distance between the
location $l$ and the center of the city $c$ (here we consider the city
of Beijing and the center is $(116.404, 39.912)$).   
\begin{equation}F_l=dist(l,c)\end{equation}
We can see from the data analysis that the distance of a location to
the center of city influences the popularity.  

\noindent{\textbf{Traffic convenience.}}
Here, we use the number of transportation stations(including bus
stations and subway stations) in $A_I$ to denote the traffic
convenience. 
\begin{equation}F_l=|\{t\in T: dist(t,l)<r\}|\end{equation}
$T$ is the set of transportation stations in Beijing. 

\noindent{\textbf{POI density.}}
Since the area of each $A_l$ for $l \in L$ is the same, the number of
POIs in $A_l$ is used to present the density of $A_l$ which is defined
as 
\begin{equation}F_l=|\{p\in P: dist(p,l)<r\}|=N(l).\end{equation}
where P is the set of all POIs in a city.

\noindent{\textbf{Popularity of specific area category.}}
We use the category of most POIs in $A_l$ to define the area category
$AC(l)=\argmax N_c(l)$. The set of all the POIs of category C is
defined as $P_C$. Then, the popularity of specific area category is
the average number of WiFi consumers for a category $C'$ over all POIs
in $P_C$ and its area category is $AC_l$. Formally,  
\begin{equation}
F_l=\frac{\sum_{p\in P'}{W(p)}}{|P'|} , \\
P'=\{p | p.c=C\cap AC(p.l)=C', p \in P\},
\end{equation}
where $p.c$ and $p.l$ are the category and location of $p$, and $W(p)$
is the number of WiFi consumers of POI $p$. 

\noindent{\textbf{Competition.}}
We also consider the competition between the stores belonging to the
same category. We define the number of POIs belonging to category C in
$A_l$ is $N_c(l)$, and the total number of POIs in $A_l$ is
$N(l)$. Then the competition is defined as the ratio of $N_c(l)$ and
$N(l)$, which is 
\begin{equation}F_l=\frac{N_c(l)}{N(l)}.\end{equation}

\noindent{\textbf{Area popularity.}}
Also, the popularity of $A_l$ can influence the popularity of $l$. The
people consumed at places in $A_l$ are the potential consumers for a
new store opened at $l$. The popularity of an area $A_l$ is defined as 
\begin{equation}F_l=\sum_{p\in A_l}W(p).\end{equation}
To avoid self-correlation of the data, we eliminate the connections of
the test POI when we evaluate the performance.  

\noindent{\textbf{Real estate price nearby.}}
The real estate price of an area reflects the economic development of
a location in the area. Thus, given a location $l$, we compute the
average price of the nearest 5 real estate in 2km to estimate the
price. 
 
\section{Evaluation}
Given a POI category or a specific brand, we first evaluate the performance of our prediction method using existing POIs belonging to the same category or brand. Then, we show some real cases of store placement using our framework.
\subsection{Metrics}
After we predict the scores for each POI, we obtain a ranked list of locations $L_P = (l_1, l_2, \cdots, l_{|L|})$. The ranked list $L_R$ of locations based on the actual popularity (number of customers) of those POIs is also obtained from the POI-Customer dataset. The position of location $l_i$ in $L_R$ is denoted with $rank(l_i)$. 

In order to evaluate the ranking performance of our result, we utilize the nDCG@k (Normalized Discounted Cumulative Gain) metric \cite{jarvelin2002cumulated} and the nSD@k (normalized symmetric difference) metric proposed in \cite{li2009consensus}.

\noindent{\textbf{nDCG@k.}}
Discounted cumulative gain (DCG) is a measure of ranking quality. DCG measures the usefulness, or gain, of an item based on its position in the result list. The gain is accumulated from the top of the result list to the bottom with the gain of each item discounted at lower ranks.
The $DCG$ at a particular rank position $p$ is defined as
\begin{equation}
DCG_p=\sum_{i=1}^p\frac{2^{rel_i}-1}{\log_2(i+1)}
\end{equation}
Sorting documents of the result list by relevance, also called Ideal DCG (IDCG) till p. The normalized discounted cumulative gain(nDCG), is computed as:
\begin{equation}
nDCG_p=\frac{DCG_p}{IDCG_p}
\end{equation}
A relevance score for an instance $l_i$ is its relative position in the actual ranking $L_R$, i.e., $rel_{l_i} = \frac{|L|-rank(li)+1 }{|L|}$.

\noindent{\textbf{nSD@k}}
The normalized symmetric difference metric provides an analysis of comparing two top-k lists.
Given two top-k lists, $\tau_1$ and $\tau_2$, the normalized symmetric difference metric is defined as:
\begin{equation}
d_{\Delta} (\tau_1, \tau_2) = \frac{1}{2k} |(\tau_1 \setminus \tau_2) \cup (\tau_2 \setminus \tau_1)|
\end{equation}
$d_{\Delta} $ value is denoted by nSD@k which is always between 0 and 1.
\begin{table}[t]
\centering
\caption{The Evaluation Data and Size}
\begin{tabular}{|c|c||c|c|} \hline
Dataset & Size & Dataset & Size \\ \hline
Coffee shops1 & 1733 &  Inn 1 & 1108\\ \hline
Coffee shops2 & 1807  & Inn 2  & 1265 \\ \hline
Starbucks & 149  &  Home-Inn  & 235 \\ \hline
Costa&  75   &  JinJiang  &57  \\ \hline
\hline\end{tabular}
\label{table:3}
\end{table}
\subsection{Evaluation data and algorithms}
\begin{figure*}[t]
\centering
\subfigure[Spatial distribution of predicted potential customers.]{ 
\label{fig:demo:a} 
\includegraphics[width=0.35\textwidth]{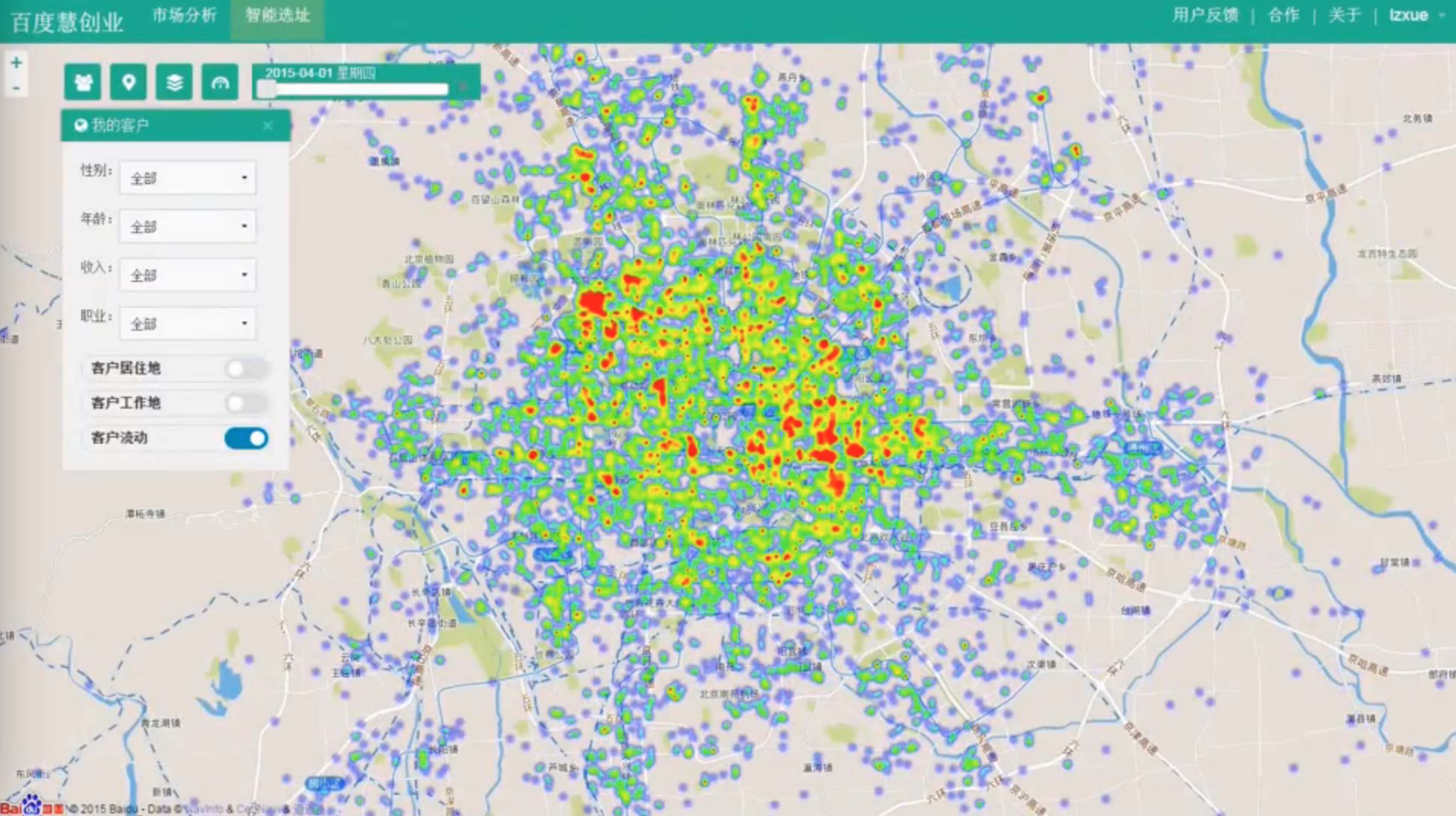}
} 
\subfigure[Location selection based on customer demands mining.]{ 
\label{fig:demo:b} 
\includegraphics[width=0.35\textwidth]{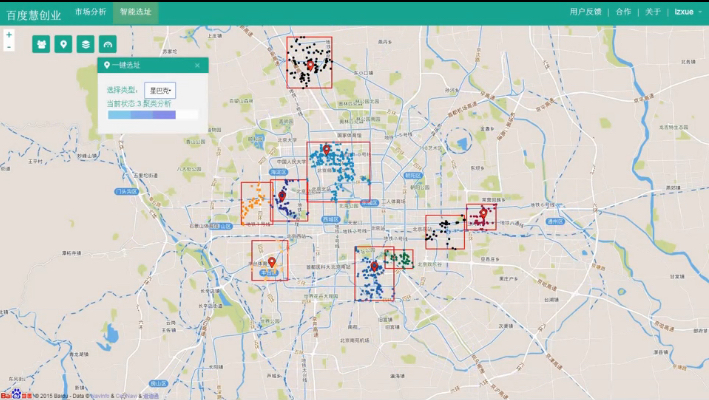}
} 
\subfigure[Location factors(competitor, transportantion, etc.) analysis.]{ 
\label{fig:demo:c} 
\includegraphics[width=0.35\textwidth]{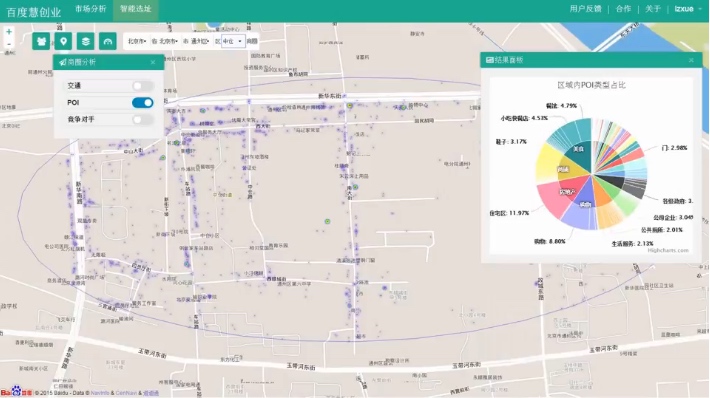}
} 
\subfigure[Candidate location comparsison analysis.]{ 
\label{fig:demo:d} 
\includegraphics[width=0.35\textwidth]{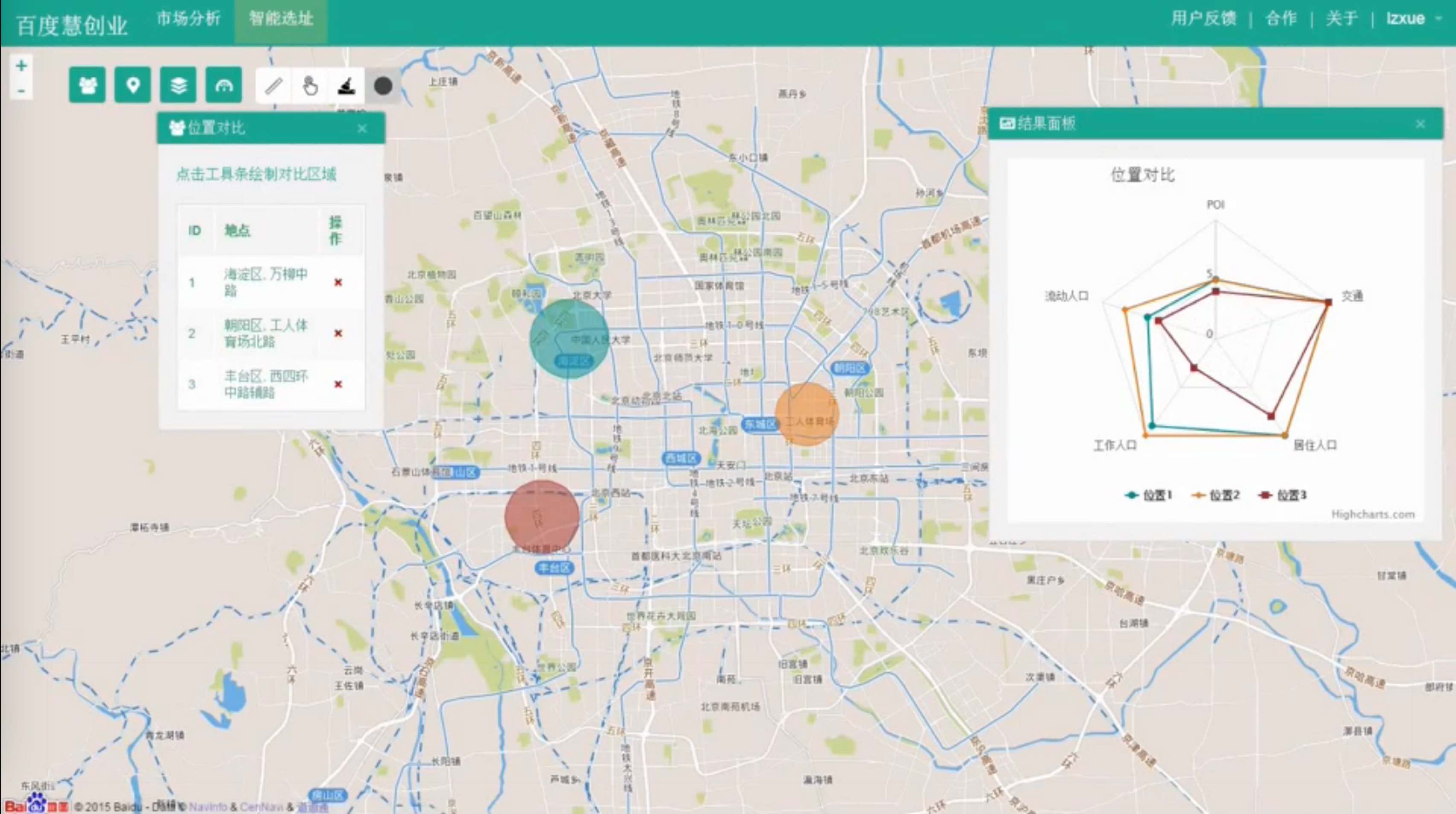}
} 
\caption{Implementation of our system. (a) shows the spatial distribution of the different kinds of users(e.g., gender, age...). (b)shows the candidate locations after analyzing user demands. (c) shows the analysis of a location including its transportation, competition and so on. (d) shows the feature comparison of several locations.}
\label{fig:demo}
\end{figure*}
We use two categories of stores which are express inns and coffee shops to evaluate the performance of our framework. For each category, the test data are two brands of stores, which is to eliminate the bias of ranking and to avoid overfitting. Specifically, for coffee shops, Starbucks and Costa are chosen as the test data, and for express inns,  two famous brands in Beijing ``Home-Inn'' and ``Jin Jiang Zhi Xing'' are chosen.  Also, the training data sets are all POI belonging to category ``coffee shop'' and ``express inn'' in Beijing except the POIs in test data. In detail, express inns exclude ``Home-Inn'' are in dataset ``Inn 1'' and those exclude ``Jin Jiang Zhi Xing'' are in dataset ``Inn 2''. In the same way, ``Coffee shops 1''  and ``Coffee shops 2'' are the two training set that exclude ``Starbucks'' and ``Costa'', respectively. The number of items of each evaluation set are shown in Table \ref{table:3}.

We have used the corresponding implementations that are public available through the Scikit-learn machine learning package \cite{scikit-learn}. The abbreviation of the evaluated algorithms and the best parameter setting after several experiments are listed below:
\begin{itemize}
\setlength{\itemsep}{0pt}
\setlength{\parsep}{0pt}
\setlength{\parskip}{0pt}
 \item  We use Linear to denote the linear regression algorithm Lasso, the regularization parameter here equals to $10^{-2}$.
 \item KernelR is the abbreviation of Kernel Regression. The kernel function of the Kernel Regression we use here is radial basis function kernel (RBF), and the regularization parameter is set to be 0.1.
 \item For SVR, the kernel function we use is also RBF. The penalty parameter of the error term is set to be 1.0, and the kernel coefficient is set to be 0.1.
 \item RF is short for Random Forests. The number of trees is 10, the minimum number of samples required to split an internal node is set to be 2. The function to measure the quality of a split is mean square error.
 \item For the Gradient Boosting for regression (GBR), the loss function to be optimized is least squares regression, and the number of boosting stages is 100.
 \item The baseline algorithm is to randomly rank the list of test set. The accuracy result is the mean of 100 times experiments.
 \item LambdaMART is a famous learn-to-rank algorithm. The number of boosting stages is 100, and the learning rate is 0.1, the minimum loss required to make a further partition is 1.0.
\end{itemize}

Also the implementation of system is shown in Figure \ref{fig:demo}. The whole system has four functions including detecting spatial distribution of different kinds of users, obtaining demand centers, location analysis and location comparison. Our framework \system\ covers two parts: obtaining demand centers and location analysis.

\subsection{Feature importance}
\begin{figure}[t]
\centering
\includegraphics[width=0.47\textwidth]{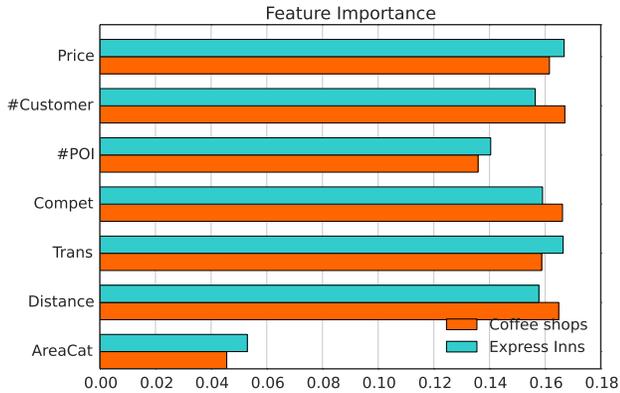}
\caption{Feature Importance for coffee shops and express inns using the gini impurity.}
\label{fig:5}
\end{figure}

Before we show the results of prediction algorithms, we use Random Forests with the gini impurity to generate the feature importance for the two categories of POIs which is shown in \ref{fig:5}. The features from top to down are the real estate price nearby, the popularity of the area, the number of POIs, the competition, the transportation, the distance to the center of the city and the category of an area. The feature importance of coffee shops and express inns are quite different. For coffee shops, the key features are the popularity of the area, the competition and the distance to the center. However, real estate price, competition and transportation are the key features for express inns. This is not surprising. The real estate price can influence the price of the express inns. When tourists choose a hotel, they tend to choose the one that is cheap and traffic convenient. Coffee shops usually attract people nearby. Competition exists everywhere and is important for all kinds of POIs. From the experimental results, we can see that the importance of features varies with the category of the store, and it's necessary to learn from the stores belonging to the same category.

\subsection{General demands accuracy}
In this subsection, we evaluate the accuracy in the situation of general demands, and the categories are the coffee shops and express inns.

We first evaluate on the test sets in Beijing using the above two metrics with different learning methods. Evaluation results in Table \ref{table:4}  show that our results using learn-to-rank algorithm and ensembled methods are far more improved with respect to the random baseline method.    

\noindent{\textbf{Accuracy with nDCG@10.}}
We measure the performance of the algorithms trained on the features we extracted.
As illustrated above, we know that the larger nDCG@10 is the more accuracy the result is. Table \ref{table:4} show the accuracy results evaluated with nDCG@10 for both coffee shops and express inns.  

With regard to the coffee shops, the best performance are using Random Forests and LambdaMART with nDCG@10 = 0.774 and nDCG@10=0.755 for ``Starbucks'' and ``Costa'', respectively. The baseline results are nDCG@10 = 0.513 and nDCG@10 = 0.523. Not only Random Forests and LambdaMART, but also the other algorithms outperform the baseline algorithm.
Learn-to-rank algorithm has larger nDCG@10 for the two test sets of express inns, whereas linear regression and baseline (randomly ranking) have lower nDCG@10. For example, in the ``Home Inn'' test, nDCG@10 for kernel regression, SVR, Random Forests, GBR and LambdaMART are 0.640, 0.680, 0.720, 0.685 and 0.736, respectively, while NDCG@10 of Linear Regression and Baseline are 0.510 and 0.540 which are less than the former ones. The JinJiang test shows the similar results, and the best accuracy is 0.767. 

\noindent{\textbf{Accuracy with nSD@10.}}
 Unlike nDCG@10 evaluates the ranking of the lists, we use nSD@10 to evaluate the distance of two top-k lists which are the top-k real list and the top-k experimental list. Note that the smaller the nSD@10 is, the more similar the two top-k lists are. Table \ref{table:4} also shows nSD@10 for both test sets of the two categories. 
 As the result of coffee shops shown,  the performances of Ensemble method algorithms are still better than the linear algorithm and the baseline algorithm. For ``Starbucks''  test, the baseline is 0.93, however, nSD@10 of Linear Regression, Kernel Regression, SVR, Random Forests, GBR and LamdaMART are 0.9, 0.9, 0.9, 0.6, 0.8 and 0.8. For ``Costa'' test, the baseline is 0.87, and the best result are Random Forests and LambdaMART with nSD@10= 0.6 which means that there are at least 4 predicted locations are in the top-10 of the real lists.  Results show that the above conclusion is also similar for the express inns. The best nSD@10 for ``Home Inn'' and ``JinJiang'' are 0.8 and 0.7.
\begin{table}
\centering
\caption{Specific demands accuracy}
\begin{tabular}{c c c c} \hline \hline
\multirow{2}{*}{Store} & \multirow{2}{*}{Algorithm} & \multicolumn{2}{c}{Measurement} \\ 
\cline{3-4} 
& & nDCG@5 & nSDK@5 \\ \hline
\multirow{7}{*}{Starbucks} & Linear & 0.530 & 0.85 \\ 
& KR & 0.624 & 0.76  \\ 
& SVR & 0.683 & 0.67 \\ 
 & RF & \textbf{0.740} & \textbf{0.56}\\ 
 & GBR & 0.727 & 0.68 \\ 
 & LambdaMART & 0.736 & \textbf{0.56}\\ 
 & Baseline & 0.510 & 0.82\\ 
 & & & \\
\multirow{7}{*}{HomeInn}& Linear & 0.555 & 0.872  \\ 
 & KR & 0.630 & 0.86\\ 
 & SVR & 0.687 & 0.75\\ 
 & RF & 0.715 & 0.65\\ 
 & GBR & 0.695 & 0.73\\ 
 & LambdaMART & \textbf{0.748} & \textbf{0.60}\\ 
 & Baseline & 0.536 & 0.892\\ \hline \hline
\end{tabular}
\label{table:5}
\end{table} 
\subsection{Specific demands accuracy}
Unlike general demands, for specific demands, we evaluate on two brands of chain coffee shops and chain express inns which are ``Starbucks'' and ``Home Inn'' using nDCG@5 and nSD@5. We randomly choose 20\% of the data as test and remaining as training data. Table \ref{table:5} shows the average results of the 10 times repeated experiments. 

Comparing with the baseline 0.510, Random Forests is the optimal learning algorithm for ``Starbucks'' with nDCG@5=0.740 which offer about 45\% improvement. For ``HomeInn'', the optimal learning algorithm is LambdaMART with nDCG@5= 0.748. Also, the lowest nSD@5 are 0.56 and 0.65 for ``Starbucks'' and ``Home Inn'' respectively. 
Above results show that our framework also works well on the chain stores.
\begin{table*}[t]
\centering
\caption{General demands accuracy}
\begin{tabular}{c c c c|c c c c} \hline \hline
\multirow{2}{*}{Store} & \multirow{2}{*}{Algorithm} & \multicolumn{2}{c|}{Measurement} & \multirow{2}{*}{Store} & \multirow{2}{*}{Algorithm} & \multicolumn{2}{c}{Measurement} \\ 
\cline{3-4} \cline{7-8}
& & nDCG@10 & nSDK@10 &
& & nDCG@10 & nSDK@10 \\ \hline
\multirow{7}{*}{Starbucks} & Linear & 0.560 & 0.90 & \multirow{7}{*}{Home-Inn} & Linear & 0.450 & 1.0 \\ 
& KR & 0.740 & 0.90 &
 & KR & 0.640 & 0.90 \\ 
& SVR & 0.650 & 0.90 &
 & SVR & 0.680 & 0.90\\ 
& RF & \textbf{0.774} & \textbf{0.60} &
 & RF & 0.720 & 0.80\\ 
& GBR & 0.750 & 0.80 &
 & GBR & 0.685 & 0.80 \\ 
& LambdaMART& 0.766 & 0.80 &
 & LambdaMART & \textbf{0.736} & \textbf{0.70}\\ 
& Baseline & 0.513 & 0.93 &
 & Baseline & 0.510 & 0.96\\ 
 & & & & & & & \\
\multirow{7}{*}{Costa}& Linear & 0.580 & 0.90  & \multirow{7}{*}{JinJiang}& Linear & 0.560 & 0.9 \\ 
& KR & 0.637 & 0.90 &
 & KR & 0.694 & 0.70\\ 
& SVR & 0.641 & 0.80 &
 & SVR & 0.683 & 0.80\\ 
& RF & \textbf{0.755} & \textbf{0.60} &
 & RF & 0.749 & \textbf{0.60}\\ 
& GBR & 0.674 & 0.70 &
 & GBR & 0.738 & 0.70\\ 
& LambdaMART& 0.746 & \textbf{0.60} &
 & LambdaMART & \textbf{0.767} & 0.70\\ 
& Baseline & 0.523 & 0.90 &
 & Baseline & 0.540 & 0.83\\ \hline \hline
\end{tabular}
\label{table:4}
\end{table*}
\begin{figure*}[t]
\centering
\subfigure[Demand centers of ``Haidilao''.]{ 
\label{fig:9:a} 
\includegraphics[width=0.4\textwidth]{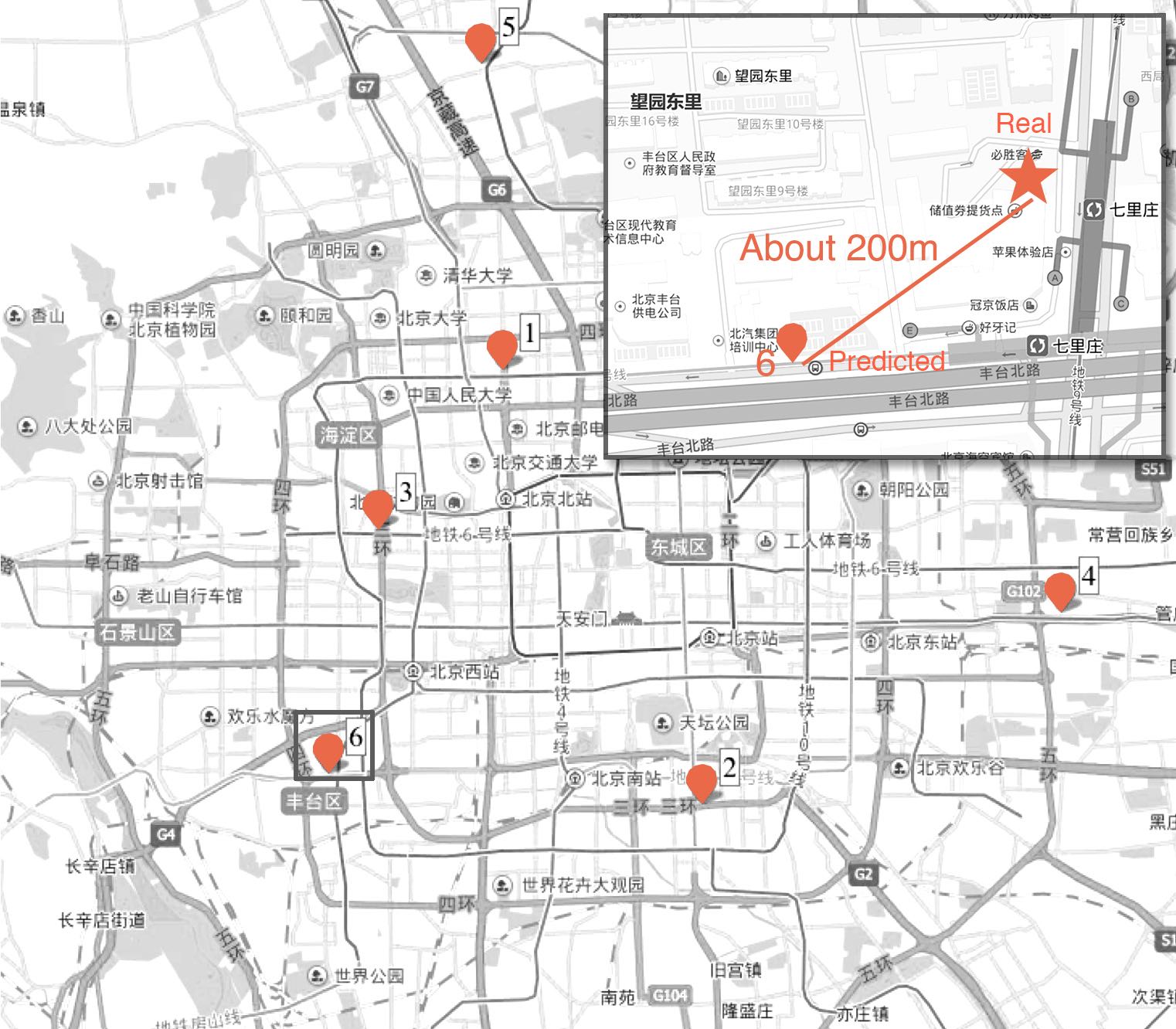}
} 
\subfigure[Demand centers of ``Starbucks''.]{ 
\label{fig:9:b} 
\includegraphics[width=0.4\textwidth]{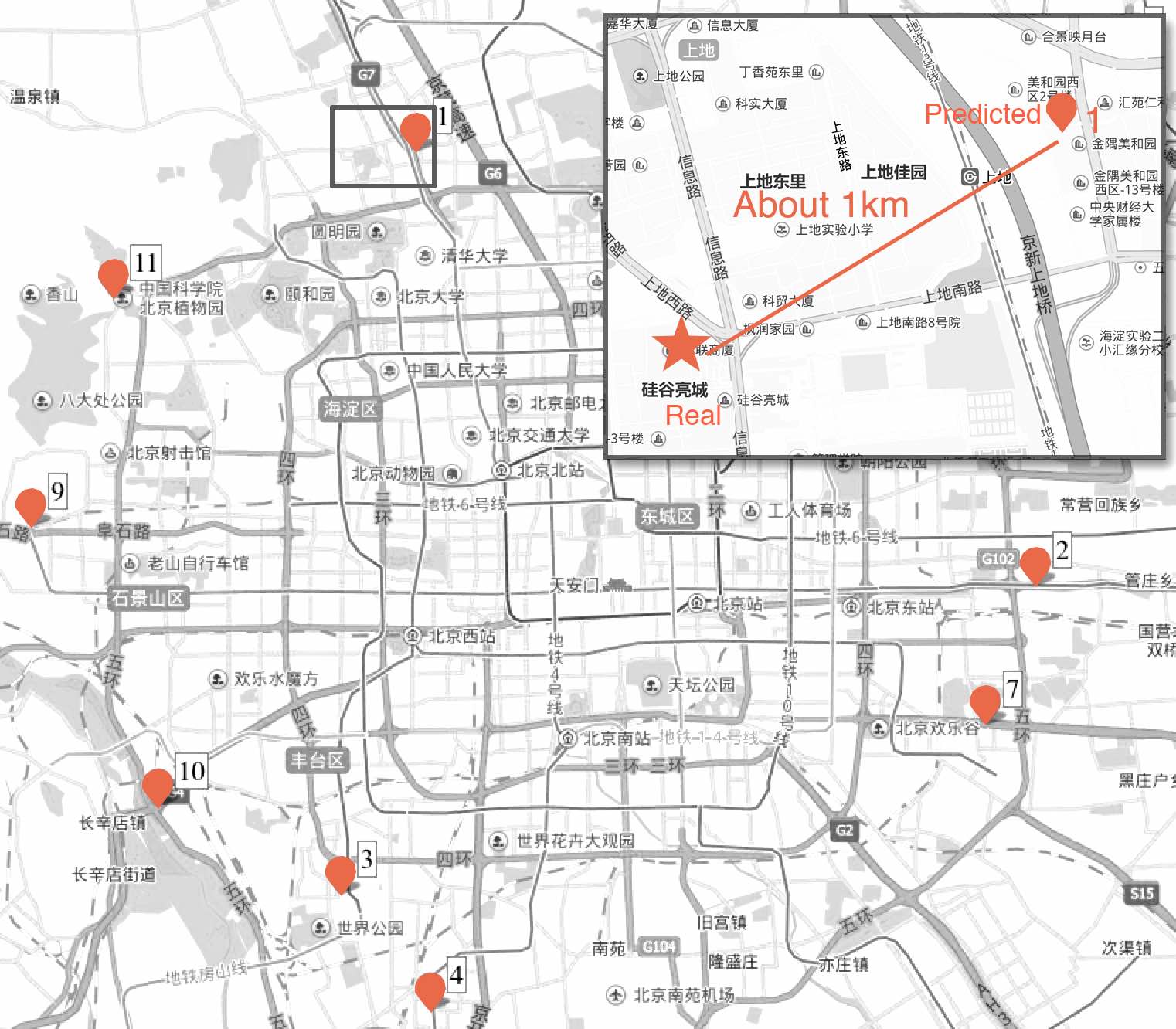}
}
\caption{Real cases. (a) and (b) show the demand centers of ``Haidilao'' and ``Starbucks'', respectively. (c) and (d) show the distance between predicted optimal placement and real placement for the two stores. The star marks denote the real placement and the circles with label are the predicted optimal locations.}
\label{fig:10}
\end{figure*}
\subsection{Real cases}
Now we study two real cases which can somehow reflect the effectiveness of our framework(both the obtaining demands part and the learning part). The two cases are ``Starbucks'' opened at January, 2016 and chain hotpot restaurant ``HaiDiLao'' opened at September, 2015. We obtain the location with demands using the map query data from April, 2015 to June, 2015 which is far early from the existence of two new stores.  The locations with demands of ``Starbucks'' and ``HaiDiLao'' are shown in Figure \ref{fig:9:a} and Figure \ref{fig:9:b}, respectively.  Note that we exclude the existing store opened at that time. We train the Random Forests model using the two brands of existing chain stores. The ranking of the locations with demands of ``HaiDiLao'' is [`4', `1', `9', `2', `6', `12', `5', `11', `8', `10', `7', `3'], and the distance between location `1' which is in the top-3 of the ranking list and the new opened store is about 1km.  For ``Starbucks'', the predicted ranking list is [`2', `6', `5', `4', `1', `3'] . Location `6' is only 200m away from the real location.
\section{Related work}
The store placement problem has be studied by researchers from various fields. Researchers have concentrated on various techniques including spatial interaction models, Multiple regression Discriminant analysis and so on \cite{hernandez2000art} . Spatial interaction models are based on the assumptions that the intensity of interaction between two locations decreases with their distance and that
the usability of a location increases with the intensity of utilization and the proximity of complementarily arranged locations 
\cite{athiyaman2010location, berman2002generalized, xiao2011optimal, kubis2007analysis}. Specifically, work in \cite{kubis2007analysis} employs spatial interaction theory, customer density estimates and minimax decision criterion to address site selection issues. Authors of 
\cite{berman2002generalized, xiao2011optimal} utilize theoretic models of optimal location by maximizing the number of residents it can attracts in Euclidean space or road network. By saying a location attracts a resident, we mean that the location is closest to the place the resident live among all the existing stores. However, in reality, people are moving and the location can attracts not only residents near it. 
Multiple regression Discriminant analysis \cite{rogers1997site} is a location analyst that has been employed to produce a series of sales forecasts for both new and existing stores, which is tested and calibrated across a number of different scenarios and is often used as a benchmarking tool for future development. Jensen \cite{jensen2006network} proposed a spatial network based formulation of the problem, where nodes are differrient types of retail stores and weighted signed links are defined to model the attraction and repulsion of entities in the network.

Location based services have been widely used in the analysis of trade and location placement \cite{karamshuk2013geo, qu2013trade}. 
Karamshuk et al. \cite{karamshuk2013geo} find optimal retail store location from a list of locations by using supervised learning with features mined from Foursquare check-in data. Researchers in \cite{li2015location} focus on locating the ambulance stations by using the real traffic information so as to minimize the average travel-time to reach the emergency requests.
The authors of \cite{qu2013trade} illustrate how User Generated Mobile Location Data like Foursquare check-ins can be used in trade area analysis.
\cite{fu2014sparse} exploits regression modeling, pairwise ranking objective and sparsity regularization to solve the real estate ranking problem with online user reviews and offline moving behaviors. Also authors in \cite{Fu:2014:EGD:2623330.2623675} propose a method for estate appraisal by leveraging the mutual enforcement of ranking and clustering power.

Previous work of finding optimal location do not consider the activity demand of users. 
 \cite{wupredict} predicts users' activity via integrating data of map query and mobility trace, which concludes the fact that map query data can be seen as demands and signals of users doing the activity. Rogers \cite{rogers2007retail} proposed some key elements of a retail location study including competition, transportation, trade area definition and so on. Inspired by the above work, we combine users' geodemand with supervised learning \cite{karamshuk2013geo} using geographic and consumptive information (some key elements) mined from multiple real data sources. 
\section{Conclusion}
In this paper, we propose a novel framework \system\ for the optimal
store placement problem. Our framework combines the spatial
distribution of user demands with the popularity and economic
attributes of each location. Also the integration of multiple data
sources enables us to extract rich features from existing stores. The
evaluation results demonstrate the effectiveness of our proposed
method for optimal store placement in both experiments and real life
business scenarios. 

\small
\balance
\bibliographystyle{abbrv}
\bibliography{location}

\end{document}